\documentclass[pdflatex,sn-nature,oneside]{sn-jnl}

\usepackage{graphicx}%
\usepackage{multirow}%
\usepackage{amsmath,amssymb,amsfonts}%
\usepackage{amsthm}%
\usepackage{mathrsfs}%
\usepackage[title]{appendix}%
\usepackage{xcolor}%
\usepackage{textcomp}%
\usepackage{manyfoot}%
\usepackage{booktabs}%
\usepackage{algorithm}%
\usepackage{algorithmicx}%
\usepackage{algpseudocode}%
\usepackage{listings}%
\usepackage[most]{tcolorbox}
\usepackage{afterpage}
\usepackage{caption}
\usepackage{url}
\usepackage{mwe}

\theoremstyle{thmstyleone}%
\theoremstyle{thmstyletwo}%

\theoremstyle{thmstylethree}%

\begin{document}

\title[Article Title]{AIBuildAI: An AI Agent for Automatically Building AI Models}

\author[1]{\fnm{Ruiyi} \sur{Zhang}}
\equalcont{These authors contributed equally to this work.}

\author[1]{\fnm{Peijia} \sur{Qin}}
\equalcont{These authors contributed equally to this work.}

\author[1]{\fnm{Qi} \sur{Cao}}
\equalcont{These authors contributed equally to this work.}

\author[1]{\fnm{Li} \sur{Zhang}}
\equalcont{These authors contributed equally to this work.}

\author*[1,2]{\fnm{Pengtao} \sur{Xie}}\email{p1xie@ucsd.edu}

\affil[1]{\orgdiv{Department of Electrical and Computer Engineering}, \orgname{University of California San Diego}, \orgaddress{\city{La Jolla}, \postcode{92093}, \state{CA}, \country{USA}}}

\affil[2]{\orgdiv{Department of Medicine}, \orgname{University of California San Diego}, \orgaddress{\city{La Jolla}, \postcode{92093}, \state{CA}, \country{USA}}}

\setlength{\oddsidemargin}{0in}
\setlength{\evensidemargin}{0in}


\abstract{AI models underpin modern intelligent systems, driving advances across science, medicine, finance, and technology. Yet developing high-performing AI models remains a labor-intensive process that requires expert practitioners to iteratively design architectures, engineer representations, implement training pipelines and refine approaches through empirical evaluation. Existing AutoML methods partially alleviate this burden but remain limited to narrow aspects such as hyperparameter optimization and model selection within predefined search spaces, leaving the full development lifecycle largely dependent on human expertise.
To address this gap, we introduce AIBuildAI, an AI agent that automatically builds AI models from a task description and training data. AIBuildAI adopts a hierarchical agent architecture in which a manager agent coordinates three specialized sub-agents: a designer for modeling strategy, a coder for implementation and debugging, and a tuner for training and performance optimization. Each sub-agent is itself a large language model (LLM) based agent capable of multi-step reasoning and tool use, enabling end-to-end automation of the AI model development process that goes beyond the scope of existing AutoML approaches.
We evaluate AIBuildAI on MLE-Bench, a benchmark of realistic Kaggle-style AI development tasks spanning visual, textual, time-series and tabular modalities. AIBuildAI ranks first on MLE-Bench with a medal rate of 63.1\%, outperforming all existing baseline methods and matching the capability of highly experienced AI engineers. These results demonstrate that hierarchical agent systems can automate the full AI model development process from task specification to deployable model, suggesting a pathway toward broadly accessible AI development with minimal human intervention.}

\maketitle

\afterpage{%
  \begin{center}
    \includegraphics[width=\linewidth]{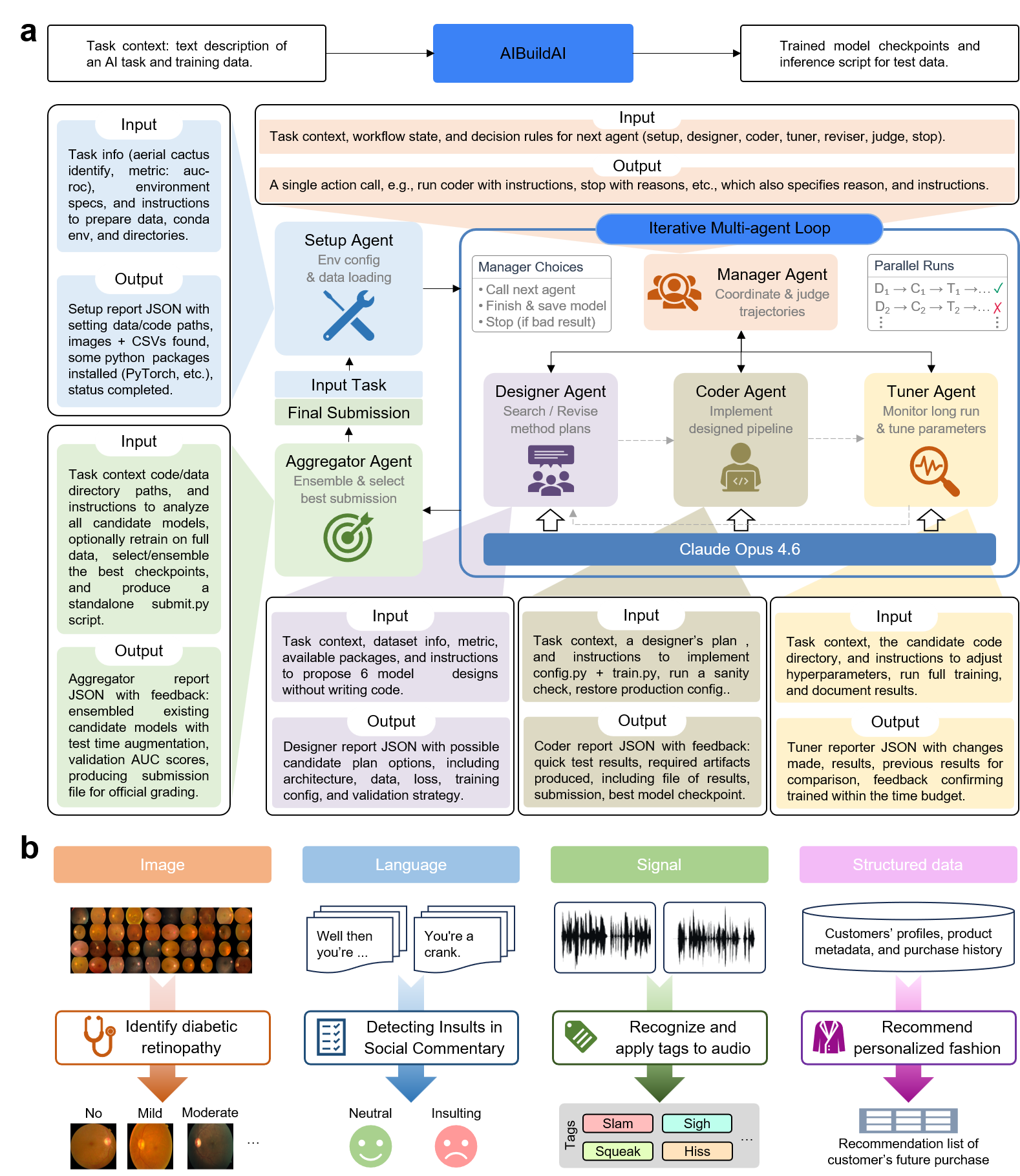}
    \vspace{1.5em}
  \captionof{figure}{\textbf{Overview of AIBuildAI.} 
\textbf{a}, AIBuildAI is an AI agent for automatically building AI models. It takes as input a text description of an AI task and training data, automatically designs and trains an AI model, and outputs the trained model checkpoints and inference script. AIBuildAI adopts a hierarchical agent architecture in which a manager agent oversees multiple parallel solutions, deciding which sub-agent to invoke for iteratively refining each solution repository. Sub-agents include a designer that proposes or revises modeling plans, a coder that implements and debugs training pipelines, and a tuner that runs full training and optimizes performance. Each sub-agent is itself an LLM-based agent that performs multiple rounds of LLM calls and tool interactions within a single invocation. Surrounding this core loop, a setup agent prepares the conda environment and installs the required packages at initialization, and an aggregator agent selects or ensembles the strongest candidate solutions at termination to produce the final AI model.
\textbf{b}, AIBuildAI is applicable to a diverse range of tasks spanning multiple data modalities and problem formulations, including visual recognition, natural language processing, temporal forecasting, and structured tabular prediction.}
  \label{fig:AIBuildAI_workflow}
  \end{center}
  \clearpage
}

\section{Introduction}\label{sec1}

\textcolor{black}{AI models underpin modern intelligent systems and are central to progress across biomedicine, chemistry, materials science, the physical sciences, and a broad range of scientific disciplines \cite{turing2007computing,jordan2015machine,ghahramani2015probabilistic,biamonte2017quantum}. Advances in computer vision, natural language processing, time-series forecasting, tabular data analysis, and related areas have enabled data-driven discovery and decision-making at scale \cite{he2016deep,otter2020survey,price2025probabilistic,hollmann2025accurate}. Despite the availability of mature learning algorithms and widely used deep learning libraries, developing high-performing AI models remains labor-intensive and reliant on specialized expertise \cite{bergstra2012random,sculley2015hidden,hutter2019automated}. The process involves iterative design of modeling strategies, data preprocessing and validation, writing code to implement these designs, building and running training pipelines, hyperparameter optimization, debugging, and continual refinement based on empirical evaluation \cite{aldoseri2023re,yang2024limits}. This complexity limits the ability of scientists and practitioners to fully leverage AI for large-scale discovery and real-world applications \cite{domingos2012few,feurer2015efficient,henderson2018deep}.}

Existing automation approaches partially alleviate this burden but remain limited in scope. Traditional AutoML methods focus on hyperparameter optimization and model selection within predefined candidate sets \cite{thornton2013auto,he2021automl}. Despite their effectiveness, AutoML methods require users to have prior knowledge to define candidates and cannot achieve full automation. More recently, large language model (LLM) based coding agents have been applied to automatically building AI models. These methods treat an AI model as a code solution and use a single LLM call to iteratively modify and search for an optimal solution~\cite{toledo2025airesearchagentsmachine,Du2025AutoMLGenNF}. However, these approaches are constrained by what can be accomplished in a single LLM call, which limits the complexity of modifications at each step. Moreover, implementation and tuning of AI models reflect two distinct phases of development with separate objectives, posing additional challenges for a single coding agent to handle.

Recent advances in LLM-based multi-agent systems have demonstrated that coordinating multiple specialized agents can effectively tackle complex tasks such as software development and scientific research~\cite{hong2024metagpt,wu2024autogen}. Building on this progress, we introduce AIBuildAI, an AI agent that automatically builds AI models from a task description and training data using multiple coordinated sub-agents. Unlike approaches based on a single coding agent, AIBuildAI adopts a hierarchical agent architecture in which a manager agent coordinates three specialized sub-agents: a designer for modeling strategy, a coder for implementation and debugging, and a tuner for training and performance optimization. Each sub-agent is itself an LLM-based agent that performs multiple rounds of LLM calls and tool interactions within a single invocation; for example, the coder sub-agent may iteratively write code, execute it, inspect errors and refine the implementation across many such rounds before returning a working solution. The manager selects which candidate solutions to develop further through long-context reasoning over the full execution history, replacing predefined search policies with adaptive, context-aware decisions.

We evaluate AIBuildAI on MLE-Bench, a benchmark of realistic Kaggle-style AI development tasks spanning visual, textual, time-series, and tabular modalities \cite{chan2025mlebench}. Without task-specific customization, AIBuildAI ranks first on the March 18, 2026 leaderboard of MLE-Bench \cite{mlebench_commit_2026} with a medal rate of 63.1\%, outperforming all existing autonomous AI development systems including AIRA-dojo~\cite{toledo2025airesearchagentsmachine} and MLEvolve~\cite{Du2025AutoMLGenNF}. These results demonstrate that hierarchical agent systems can automate the full AI model development process, from task specification to deployable model, matching the capability of highly experienced AI engineers with minimal human intervention.

\section{Results}\label{Results}

\subsection{Overview of AIBuildAI}

AIBuildAI is an AI agent for automatically building AI models. It takes as input a text description of an AI task and training data, automatically designs, implements and trains an AI model with high performance, and outputs the trained model checkpoints and inference script that can be directly applied to unseen test data. AIBuildAI adopts a hierarchical agent architecture inspired by real-world AI research processes. In traditional AI projects, a technical lead oversees several parallel lines of investigation, each exploring a different modeling direction. Within each line, researchers design modeling strategies while engineers implement training pipelines and run experiments to obtain empirical results. The lead periodically reviews the outcomes of these experiments, allocates resources to promising directions, requests revisions when progress stalls, and terminates unproductive efforts to focus on stronger candidates. AIBuildAI mirrors this process through a structured, agent-driven workflow (Fig.~\ref{fig:AIBuildAI_workflow}a).

The core workflow of AIBuildAI operates over multiple parallel solution candidates, each maintained as an isolated workspace (repository) containing the artifacts that define an AI modeling pipeline, including the model design description, training code, configuration and hyperparameter files, logs, and evaluation results.
Solution repositories are initialized empty and iteratively refined until the time budget is exhausted, under a trio of core sub-agents (designer, coder, and tuner) orchestrated by a manager agent. Each sub-agent is capable of multi-step reasoning and tool usage through multiple LLM calls within a single agent invocation.
The designer sub-agent is responsible for modeling and training strategy, proposing new approaches informed by web search or revising existing plans based on training and evaluation outcomes. The coder sub-agent focuses on producing correct code from a textual design description, with operations that include reading the design, translating it into executable training and inference pipelines, writing and running code, inspecting errors, refining the implementation, and performing iterative debugging. The tuner sub-agent trains the model and tunes parameters to maximize performance, with operations that include launching training runs, monitoring metrics such as losses, reasoning over logs, adjusting hyperparameters, and iterating.
A manager agent orchestrates this process by monitoring the current status across solution repositories and deciding which sub-agent to invoke on each. This requires long-context reasoning over the accumulated execution history, adaptively composing different sequences of operations depending on the needs of each solution. In principle, promising solutions are further refined through targeted updates, while underperforming ones are discarded by ceasing further actions.
Surrounding this core loop are two lightweight components: a setup agent that sets up the conda environment with the necessary packages, and an aggregator agent that selects or ensembles the strongest candidates to produce the final submission-ready model.

\afterpage{%
  \begin{center}
    \includegraphics[width=.92\linewidth]{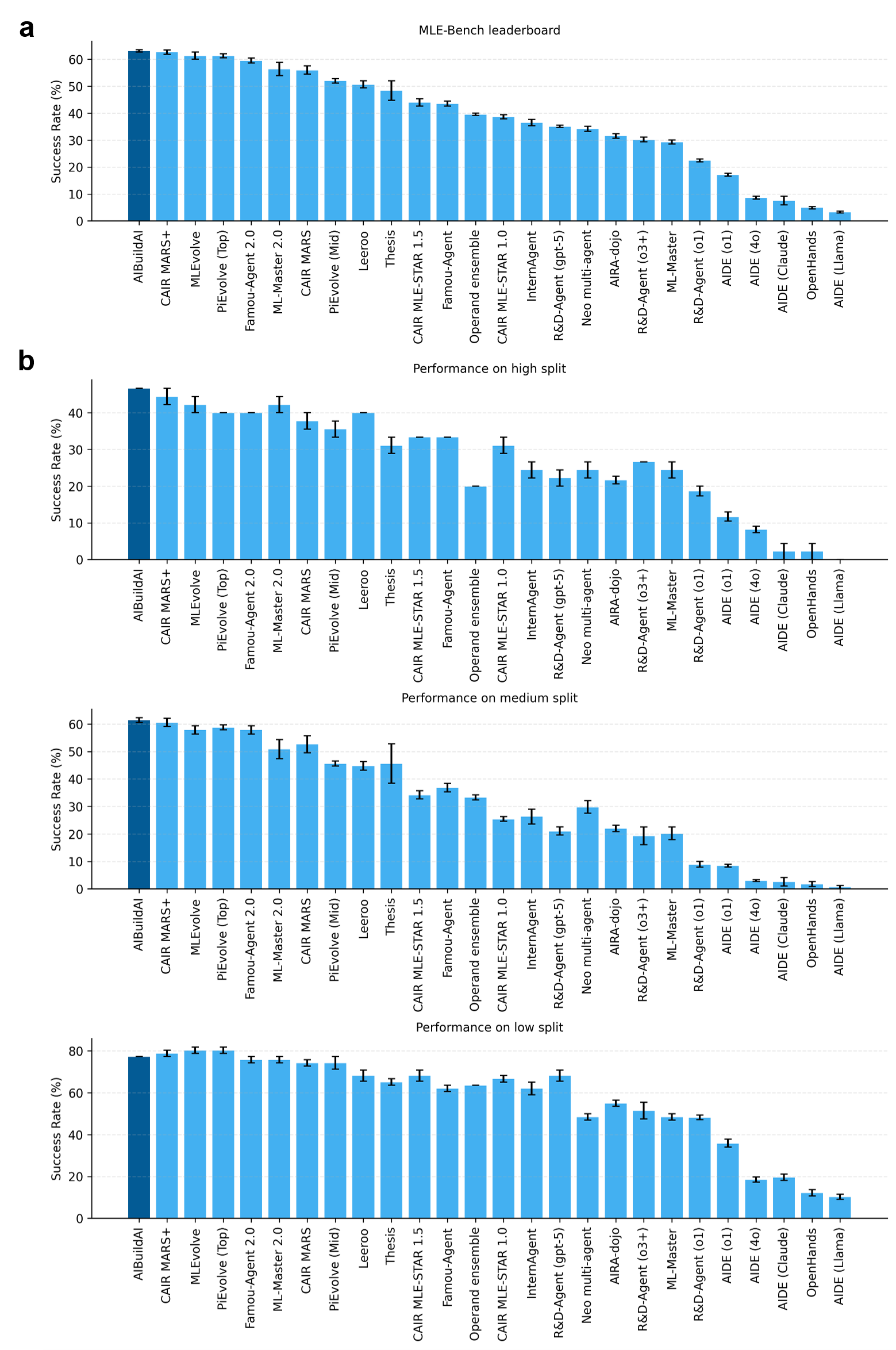}
  \captionof{figure}{\textbf{AIBuildAI ranks first on the MLE-Bench leaderboard.}
\textbf{a}, Overall medal rate of AIBuildAI compared with 26 recent baseline methods (or method variants) on MLE-Bench as of the March 18, 2026 leaderboard. AIBuildAI achieves the top overall position with a medal rate of 63.1\%, surpassing prior systems including MARS, Famou-Agent, ML-Master, Leeroo, InternAgent, R\&D-Agent, AIDE, AIRA-dojo, and MLEvolve.
\textbf{b}, Medal rates of AIBuildAI and baseline methods on the three complexity splits of MLE-Bench. AIBuildAI ranks first on both the medium- and high-complexity splits, demonstrating that its performance advantage becomes more pronounced on more challenging AI tasks.}
  \label{fig:mle-bench}
  \end{center}
  \clearpage
}

AIBuildAI is applicable to a diverse range of AI tasks across multiple data modalities and problem formulations, including visual recognition, natural language processing, temporal forecasting, and structured tabular prediction (Fig.~\ref{fig:AIBuildAI_workflow}b). In the following sections, we first evaluate AIBuildAI on MLE-Bench against state-of-the-art baseline methods, and then compare its detailed task performance against two representative baselines, AIRA-dojo~\cite{toledo2025airesearchagentsmachine} and MLEvolve~\cite{Du2025AutoMLGenNF}, across 10 image classification tasks, 10 detection, segmentation, and video understanding tasks, 10 language understanding and generation tasks, and 6 temporal and tabular tasks.

\subsection{AIBuildAI ranks first on the MLE-Bench leaderboard}

We evaluate AIBuildAI on MLE-Bench, one of the most comprehensive benchmarks for assessing autonomous AI systems on realistic Kaggle-style tasks that require end-to-end AI model development rather than isolated algorithmic components~\cite{chan2025mlebench}. MLE-Bench contains 75 tasks curated from past Kaggle competitions, each providing raw datasets, evaluation metrics, and submission protocols faithful to real-world AI development, and requiring systems to design modeling strategies, implement training pipelines, and iteratively improve performance under limited computational budgets. Performance is measured using the medal system derived from human Kaggle competition rankings, where submissions that meet predefined percentile thresholds on the Kaggle leaderboard are awarded medals. As Kaggle competitions draw a global community of expert AI researchers and developers, earning a medal serves as a direct proxy for AI development capability on par with top human practitioners.

AIBuildAI ranks first on the MLE-Bench leaderboard with a medal rate of 63.1\% as of March 18, 2026 (Fig.~\ref{fig:mle-bench}a), establishing a new state of the art for autonomous AI development. AIBuildAI outperforms all 26 recent baseline methods (or method variants) on the leaderboard, including MARS~\cite{chen2026mars}, Famou-Agent~\cite{li2025fmagent}, ML-Master~\cite{liu2025mlmaster}, Leeroo~\cite{nadaf2026kapso}, InternAgent~\cite{team2025internagent}, R\&D-Agent~\cite{yang2025rdagent}, AIDE~\cite{jiang2025aide}, AIRA-dojo~\cite{toledo2025airesearchagentsmachine}, and MLEvolve~\cite{Du2025AutoMLGenNF}. Together, these results demonstrate that AIBuildAI matches the capability of top-tier human AI researchers and developers across a broad spectrum of real-world AI development tasks without any task-specific customization, highlighting the effectiveness of its hierarchical sub-agent design for autonomous AI model development.

Fig.~\ref{fig:mle-bench}b further reports the performance of AIBuildAI across the three complexity splits of MLE-Bench, comprising 22 low-complexity, 38 medium-complexity, and 15 high-complexity tasks, with complexity reflecting the difficulty of building a competitive solution. AIBuildAI achieves medal rates of 77.27\%, 61.40\%, and 46.67\% on the low-, medium-, and high-complexity splits, respectively, ranking first on both the medium- and high-complexity splits. The high-complexity split represents the most challenging and practically important tier of MLE-Bench, demanding multi-step strategy design, careful implementation, and extensive tuning to obtain a competitive solution. The strong performance of AIBuildAI on this tier underscores its advantage in building AI models for the most demanding tasks.

\afterpage{%
  \begin{center}
    \includegraphics[width=\linewidth]{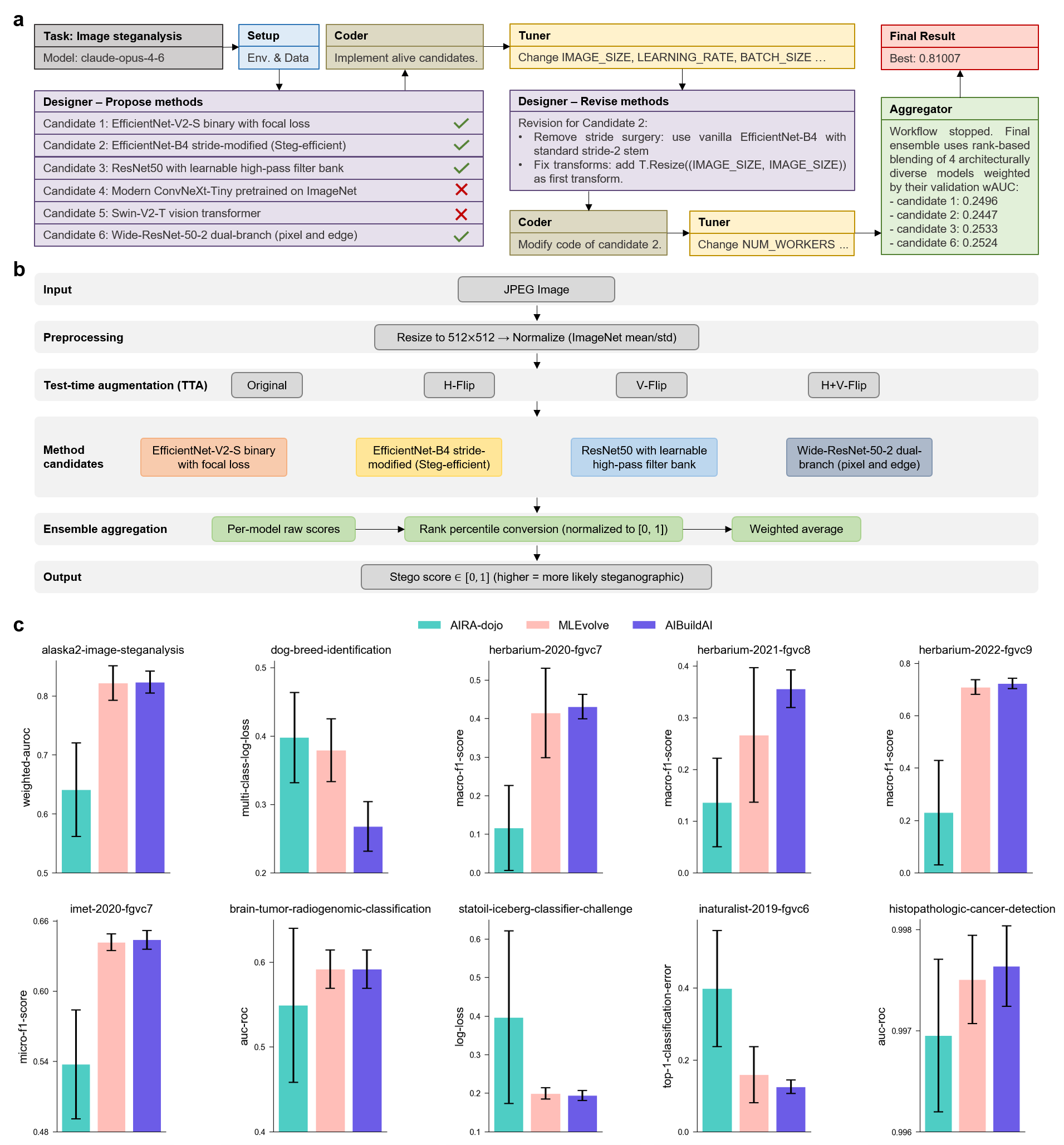}
  \captionof{figure}{\textbf{AIBuildAI outperforms baseline methods on image classification tasks.} 
\textbf{a,} Example workflow on the alaska2-image-steganalysis task, a steganalysis challenge to predict the likelihood that an image contains a hidden message. The manager agent invokes the designer sub-agent to propose six solution candidates based on diverse architectures including EfficientNet, ResNet, ConvNeXt, Swin Transformer, and Wide-ResNet variants. The manager reviews the six plans, selects four to proceed, and invokes the coder sub-agent to implement each plan and the tuner sub-agent to optimize training parameters and obtain validation results. Based on these results, the manager invokes the designer to revise candidate 2 by removing stride surgery and reverting to vanilla EfficientNet-B4, which then passes through the coder and tuner again. The aggregator constructs a rank-based ensemble of the four models, achieving a score of 0.8101 on the private test set.
\textbf{b,} Final solution in detail. The pipeline takes an image as input, resizes and normalizes it using ImageNet statistics. Test-time augmentation produces four views (original, H-flip, V-flip, H+V-flip). Each view is passed through four ensemble models, namely EfficientNet-V2-S with focal loss, EfficientNet-B4, ResNet50 with a learnable high-pass filter bank, and Wide-ResNet-50-2 with a dual-branch architecture. Per-model raw scores are converted to rank percentiles and combined via weighted averaging to produce a final stego score.
\textbf{c,} Performance across 10 image classification tasks spanning natural image recognition, fine-grained species classification, medical imaging, data security, and object recognition. AIBuildAI consistently outperforms baseline systems AIRA-dojo and MLEvolve across various visual domains and evaluation metrics.}
  \label{fig:image-cls}
  \end{center}
  \clearpage
}

\subsection{AIBuildAI demonstrates strong performance on image classification tasks}

Image classification involves predicting semantic category labels from visual inputs \cite{imagenet2017,he2016deep}, commonly addressed by fine-tuning pretrained convolutional neural networks such as ResNet~\cite{he2016deep} and EfficientNet~\cite{tan2019efficientnet}, or vision transformers such as ViT~\cite{dosovitskiy2021an} and Swin Transformer~\cite{liu2021swin}, potentially with data augmentation strategies \cite{wang2018deep,AutoAugment2019}. We evaluate AIBuildAI on 10 image classification tasks encompassing natural image recognition (dog-breed-identification), fine-grained species classification (herbarium-2021-fgvc8, herbarium-2022-fgvc9, herbarium-2025-fgvc7, inaturalist-2019-fgvc6, imet-2020-fgvc7), medical imaging classification (brain-tumor-radiogenomic-classification, histopathologic-cancer-detection), data-security-related image classification (alaska2-image-steganalysis), and object recognition (statoil-iceberg-classifier-challenge). As shown in Fig.~\ref{fig:image-cls}c, AIBuildAI outperforms both baselines (MLEvolve and AIRA-dojo) on all 10 tasks, demonstrating its strong capability in solving image classification problems across diverse visual domains.

Fig.~\ref{fig:image-cls}a illustrates an example workflow of AIBuildAI on an image classification task: alaska2-image-steganalysis. It is a steganalysis challenge where the goal is to predict the likelihood that a given image contains a secret message embedded using one of several steganographic algorithms. To address this task, the AIBuildAI manager agent invokes the designer sub-agent to propose six solution candidates based on diverse architectures including EfficientNet~\cite{tan2019efficientnet}, ResNet~\cite{he2016deep}, ConvNeXt~\cite{liu2022convnext}, Swin Transformer~\cite{liu2021swin}, and Wide-ResNet~\cite{zagoruyko2016wide} variants. The manager reviews the six plans and selects four to proceed, invoking the coder sub-agent to implement each plan and the tuner sub-agent to optimize training parameters and obtain validation results. Based on these results, the manager invokes the designer sub-agent to revise candidate 2 by removing stride surgery and reverting to vanilla EfficientNet-B4, which then passes through the coder and tuner again. Finally, the aggregator agent constructs a rank-based ensemble of the four models, achieving a score of 0.8101 on the private test set.

Fig.~\ref{fig:image-cls}b depicts the final solution of AIBuildAI on the alaska2-image-steganalysis task in detail. The pipeline takes an image as input, resizes and normalizes it using ImageNet~\cite{deng2009imagenet} statistics. Test-time augmentation is applied via horizontal flips, vertical flips, and the combination of both, producing four views per image (original, H-flip, V-flip, H+V-flip). Each view is independently passed through the four ensemble models, namely EfficientNet-V2-S~\cite{tan2021efficientnetv2} with focal loss~\cite{Lin2017FocalLF}, EfficientNet-B4~\cite{tan2019efficientnet}, ResNet50~\cite{he2016deep} with a learnable high-pass filter bank, and Wide-ResNet-50-2~\cite{zagoruyko2016wide} with a dual-branch pixel-and-edge architecture. Each model produces per-image raw scores, which are then converted to rank percentiles normalized to $[0, 1]$ across the test set. The rank percentiles from the four models are combined via weighted averaging to produce a final stego score between 0 and 1, where higher values indicate a greater likelihood of steganographic content.

\afterpage{%
  \begin{center}
    \includegraphics[width=\linewidth]{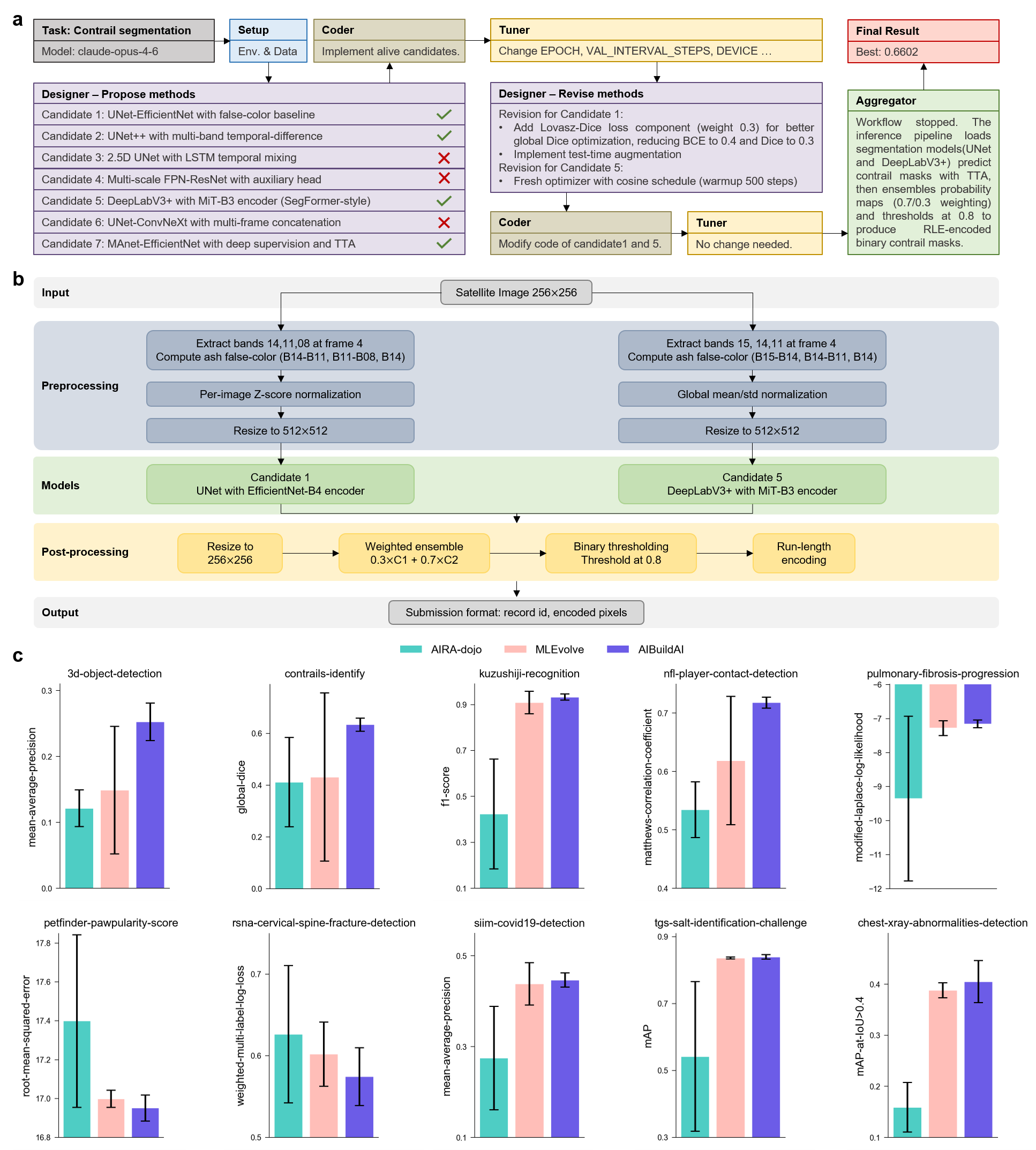}
  \captionof{figure}{\textbf{AIBuildAI surpasses baseline methods on detection, segmentation, and video understanding tasks.} 
\textbf{a,} Example workflow on the contrails-identify task, which aims to generate masks of aircraft condensation trails from satellite imagery. The manager agent invokes the designer sub-agent to propose seven solution candidates based on diverse segmentation architectures including UNet, FPN-ResNet, DeepLabV3+, and ConvNeXt variants. The manager reviews the plans, selects four candidates to proceed with, and invokes the coder and tuner sub-agents to implement, train and evaluate them. Based on the results, the manager invokes the designer to revise candidates 1 and 5 by adding a Lov\'{a}sz-Dice loss component and a fresh optimizer with cosine schedule, respectively. The coder and tuner implement and retrain the revised models. The aggregator constructs a weighted ensemble of the two strongest candidates, achieving a Dice score of 0.660 on the private test set.
\textbf{b,} Final solution in detail. The pipeline takes a satellite image and processes it through two parallel branches. The UNet branch (EfficientNet-B4 encoder) computes ash false-color composites from selected satellite bands with per-image Z-score normalization. The DeepLabV3+ branch uses a MiT-B3 encoder, a hierarchical vision transformer, with a different set of bands and global normalization. Each model produces a per-pixel probability map indicating contrail presence. The two maps are combined with weighted averaging, thresholded, and encoded in run-length format.
\textbf{c,} Performance across 10 visual perception tasks spanning object detection, semantic segmentation, medical imaging, character recognition, regression, and video-based activity recognition. AIBuildAI consistently outperforms baseline systems AIRA-dojo and MLEvolve across diverse spatial and temporal modalities.}
  \label{fig:image-others}
  \end{center}
  \clearpage
}

\subsection{AIBuildAI excels in detection, segmentation, and video understanding tasks}

Detection, segmentation, and video understanding involve predicting spatially or temporally structured outputs from visual inputs \cite{liu2020survey,sharma2022survey}, typically addressed using architectures such as Faster R-CNN~\cite{ren2015faster} and YOLO~\cite{redmon2016yolo} for object detection, UNet~\cite{ronneberger2015unet} and DeepLabV3+~\cite{chen2017deeplabv3} for semantic segmentation, and two-stream networks~\cite{simonyan2014twostream} or video transformers~\cite{arnab2021vivit} for temporal modeling. We evaluate AIBuildAI on 10 visual perception tasks involving structured prediction from images or videos, including object detection (3d-object-detection), semantic segmentation (contrails-identify, tgs-salt-identification-challenge), medical imaging (rsna-cervical-spine-fracture-detection, siim-covid19-detection, chest-xray-abnormalities-detection, pulmonary-fibrosis-progression), character recognition (kuzushiji-recognition), pet scoring regression (petfinder-pawpularity-score), and video-based activity recognition (nfl-player-contact-detection). As shown in Fig.~\ref{fig:image-others}c, AIBuildAI outperforms both baselines (MLEvolve and AIRA-dojo) on all 10 tasks, highlighting its effectiveness in handling spatially and temporally structured visual prediction tasks.

Fig.~\ref{fig:image-others}a illustrates an example workflow of AIBuildAI on a semantic segmentation task: contrails-identify. It aims to generate semantic masks of contrails (condensation trails formed by aircraft engines) from satellite images. To address this task, the AIBuildAI manager agent invokes the designer sub-agent to propose seven solution candidates based on diverse segmentation architectures including UNet~\cite{ronneberger2015unet}, FPN-ResNet~\cite{he2016deep}, DeepLabV3+~\cite{chen2017deeplabv3}, and ConvNeXt~\cite{liu2022convnext} variants. The manager reviews the seven plans, selects a subset of four to proceed, and invokes the coder and tuner sub-agents to implement and evaluate them. Based on the initial results, the manager invokes the designer sub-agent to revise candidates 1 and 5: for candidate 1, it adds a Lov\'{a}sz-Dice loss~\cite{berman2018lovasz,milletari2016vnet} component and implements test-time augmentation; for candidate 5, it introduces a cosine learning rate scheduler for the optimizer. The coder sub-agent implements the revised designs and the tuner sub-agent retrains the updated models. Finally, the aggregator agent constructs a weighted ensemble of the two strongest candidates, achieving a Dice score of 0.660 on the private test set.

Fig.~\ref{fig:image-others}b depicts the final solution of AIBuildAI on the contrails-identify task in detail. The pipeline takes a satellite image as input and feeds it through two parallel branches. The first branch uses a UNet~\cite{ronneberger2015unet} with an EfficientNet-B4~\cite{tan2019efficientnet} encoder. It selects specific satellite spectral bands and combines them into a false-color image that highlights contrails, applies per-image Z-score normalization, and upsamples the input to a higher resolution for segmentation. The second branch uses a DeepLabV3+~\cite{chen2017deeplabv3} with a SegFormer encoder (MiT-B3)~\cite{xie2021segformer} fine-tuned on ImageNet-1k~\cite{deng2009imagenet}, using a different set of satellite bands with global standardization. Each model produces a per-pixel probability map indicating contrail presence, which is upsampled back to the original input resolution. The two probability maps are then combined with a 0.3/0.7 weighting (UNet/DeepLabV3+) and thresholded at 0.8 to produce binary contrail masks.

\afterpage{%
  \begin{center}
    \includegraphics[width=\linewidth]{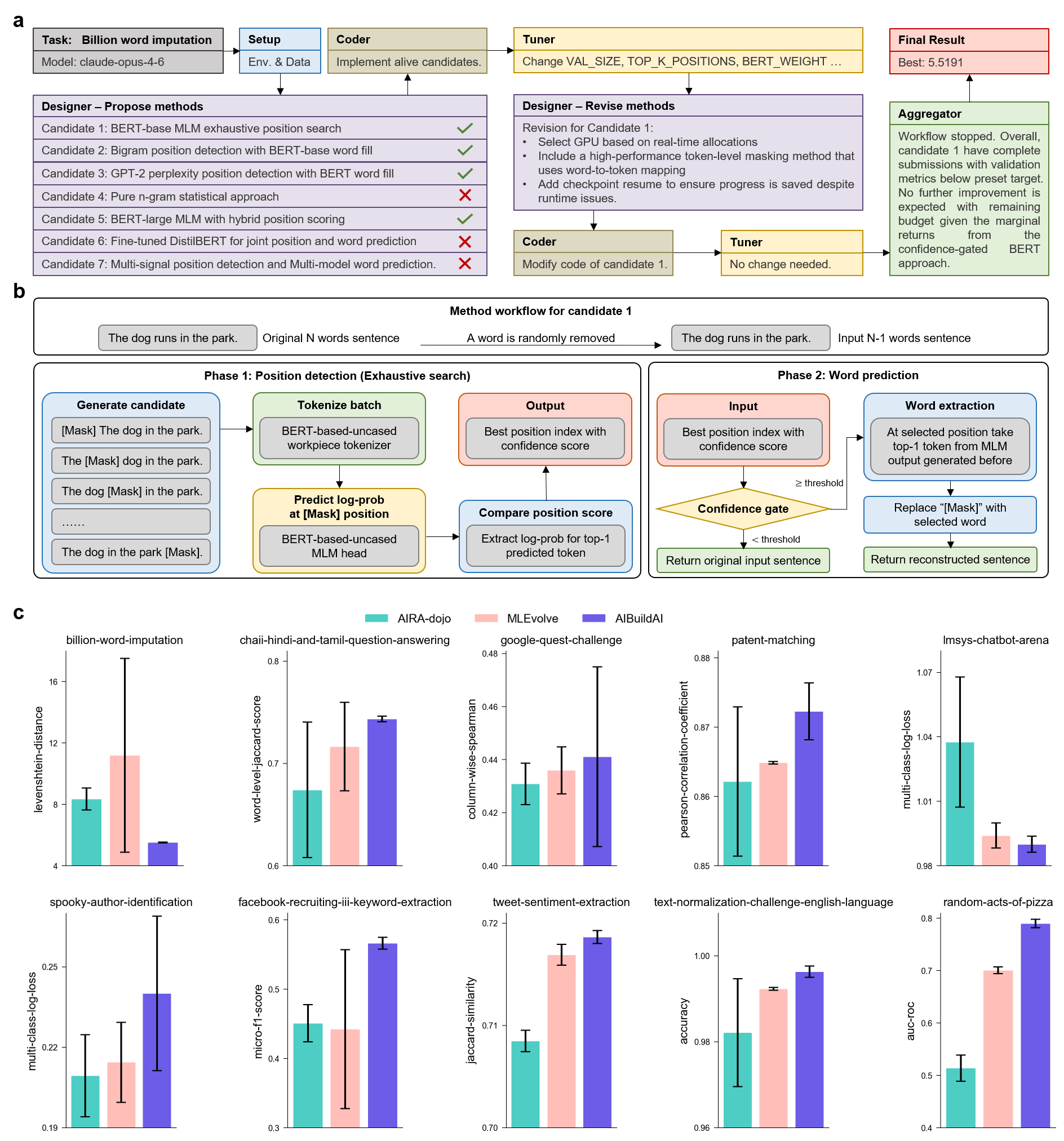}
  \captionof{figure}{\textbf{AIBuildAI outperforms baseline methods on language understanding and generation tasks.} 
\textbf{a,} Example workflow on the billion-word-imputation task, which aims to predict a missing word at an unknown position in a text paragraph. The manager agent invokes the designer sub-agent to propose seven solution candidates based on diverse approaches including BERT, DistilBERT, n-gram, and multi-model methods. The manager reviews the plans, selects a subset to proceed, and invokes the coder and tuner sub-agents to implement and evaluate them. Based on the results, the manager invokes the designer to revise candidate 1 by incorporating a token-level masking method with word-to-token mapping. The coder and tuner implement and retrain the revised model. The aggregator selects candidate 1 based on validation metrics, achieving a Levenshtein distance of 5.519 on the private test set.
\textbf{b,} Final solution in detail. Given an input sentence with a missing word, the pipeline generates candidates by inserting a mask token at every possible position. Each candidate is tokenized and passed through a BERT-base-uncased model, which computes log-probabilities over the vocabulary at the masked position. The position with the highest predicted-word confidence is selected. A confidence gate then determines whether to replace the mask with the predicted word and return the reconstructed sentence, or retain the original input unchanged.
\textbf{c,} Performance across 10 language-related tasks spanning text classification, question answering, information extraction, sequence-to-sequence generation, word imputation, and sentiment analysis. AIBuildAI consistently outperforms baseline systems, demonstrating strong capability across diverse language understanding and generation tasks.}
  \label{fig:text}
  \end{center}
  \clearpage
}

\subsection{AIBuildAI achieves superior performance on language understanding and generation tasks}

Language understanding and generation require modeling textual data for both classification and sequence prediction objectives \cite{bengio2003neural}, with standard approaches relying on fine-tuning pretrained language models such as BERT~\cite{devlin2019bert} and RoBERTa~\cite{liu2019roberta} for classification and extraction tasks, and sequence-to-sequence models such as T5~\cite{Raffel2020t5} and GPT~\cite{radford2019gpt2} for generation tasks. We evaluate AIBuildAI on 10 language-related tasks, covering text classification (patent-matching, lmsys-chatbot-arena, spooky-author-identification), question answering (chaii-hindi-and-tamil-question-answering, google-quest-challenge), information extraction (facebook-recruiting-iii-keyword-extraction), sequence-to-sequence generation (text-normalization-challenge-english-language), word imputation (billion-word-imputation), sentiment analysis (tweet-sentiment-extraction), and text-based prediction (random-acts-of-pizza). As shown in Fig.~\ref{fig:text}c, AIBuildAI outperforms both baselines (MLEvolve and AIRA-dojo) on all 10 tasks, demonstrating its effectiveness in solving diverse language understanding and generation problems.

Fig.~\ref{fig:text}a illustrates an example workflow of AIBuildAI on a language understanding task: billion-word-imputation. It aims to predict a missing word at an unknown position in a given text paragraph. To address this task, the AIBuildAI manager agent invokes the designer sub-agent to propose seven solution candidates based on diverse approaches including BERT-base MLM~\cite{devlin2019bert}, BERT-large, DistilBERT~\cite{sanh2019distilbert}, n-gram statistical methods, and multi-model ensembles. The manager reviews the seven plans, selects a subset of four to proceed, and invokes the coder and tuner sub-agents to implement, train and evaluate them. Based on the preliminary results, the manager invokes the designer sub-agent to revise candidate 1 by incorporating a token-level masking method with word-to-token mapping. The coder sub-agent implements the revised design and the tuner sub-agent retrains the model. Finally, the aggregator agent evaluates the completed submissions and selects candidate 1 based on validation metrics, achieving a Levenshtein distance of 5.519 on the private test set.

Fig.~\ref{fig:text}b depicts the final solution of AIBuildAI on the billion-word-imputation task in detail. Given an input sentence with a missing word at an unknown position, the pipeline generates candidate sentences by inserting a mask token at every possible position. Each candidate is tokenized using a wordpiece tokenizer and fed into a BERT-base-uncased model~\cite{devlin2019bert}, which computes the log-probability over the vocabulary at the masked position and extracts the score of the most likely predicted word. The position scores across all candidates are compared, and the position with the highest confidence is selected. A confidence gate then determines whether the prediction is reliable: if the confidence exceeds a predefined threshold, the mask is replaced with the predicted word to produce the reconstructed sentence; otherwise, the original input sentence is returned unchanged.

\afterpage{%
  \begin{center}
    \includegraphics[width=\linewidth]{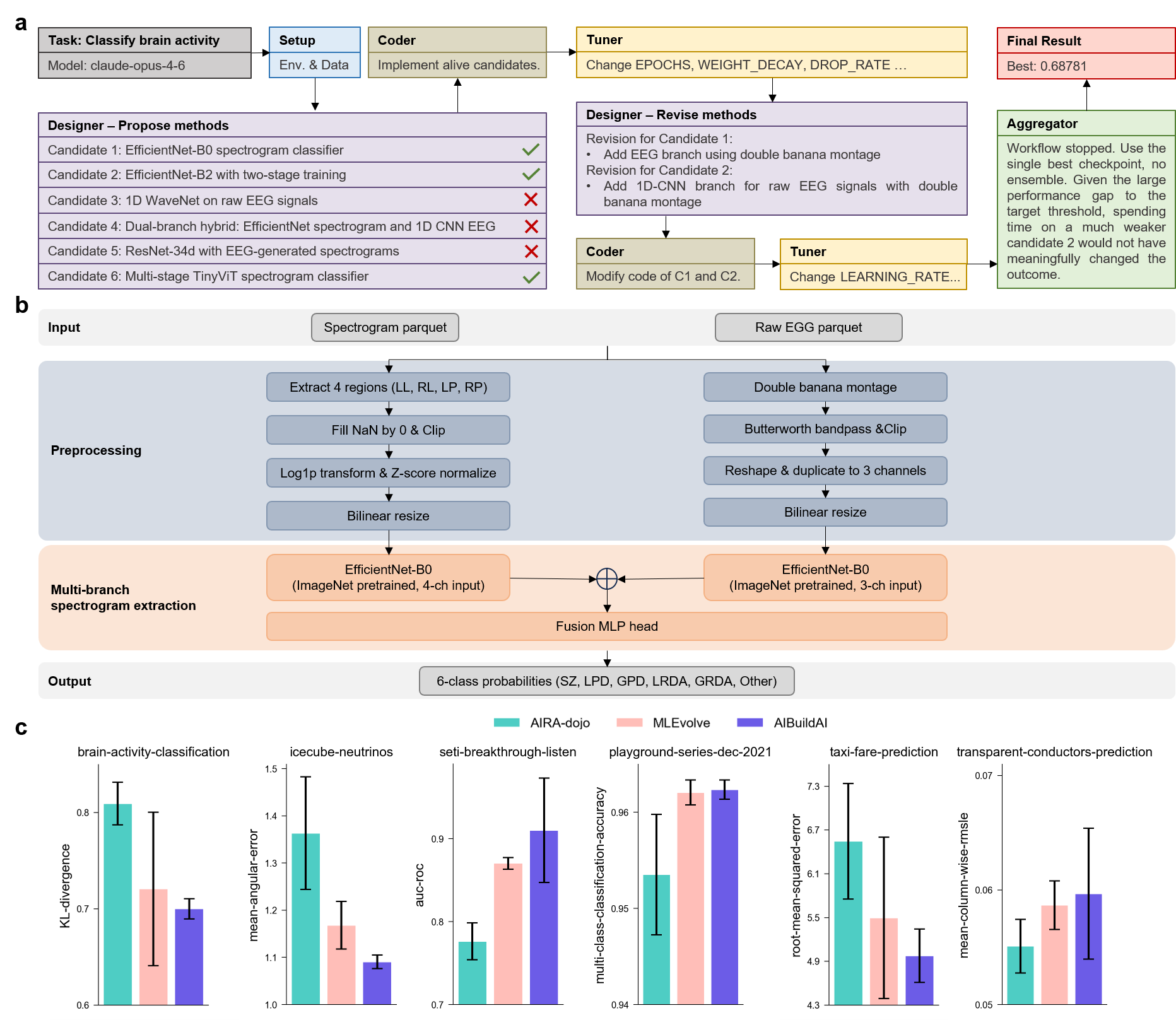}
  \captionof{figure}{\textbf{AIBuildAI surpasses baseline methods on temporal and tabular tasks.} 
\textbf{a,} Example workflow on the hms-harmful-brain-activity-classification task, which requires classifying EEG recordings into six categories of harmful brain activity (e.g., seizure, lateralized periodic discharges). The manager agent invokes the designer sub-agent to propose six solution candidates based on diverse approaches including EfficientNet spectrogram classifiers, WaveNet on raw EEG, ResNet, TinyViT, and dual-branch hybrids. The manager reviews the plans, selects a subset to proceed, and invokes the coder and tuner sub-agents to implement and evaluate them. Based on the results, the manager invokes the designer to revise candidates 1 and 2 by adding an EEG branch with double banana montage and a 1D-CNN branch for raw signals, respectively. The coder and tuner implement and retrain the revised models. The aggregator selects the best checkpoint from candidate 1 as the final model.
\textbf{b,} Final solution in detail. The pipeline takes spectrogram and raw EEG inputs and processes them through a dual-branch architecture. The spectrogram branch extracts four EEG regions (e.g., left lateral, right lateral), fills missing values, applies log-transform with Z-score normalization, and feeds the result into an ImageNet-pretrained EfficientNet-B0. The EEG branch applies a double banana montage and Butterworth bandpass filtering, then passes the processed signal through a separate EfficientNet-B0. A fusion MLP head combines both branches to produce 6-class probabilities.
\textbf{c,} Performance across 3 temporal signal modeling tasks (EEG classification, neutrino detection, radio signal search) and 3 structured tabular prediction tasks (regression, materials property prediction, general tabular prediction). AIBuildAI consistently outperforms baseline systems AIRA-dojo and MLEvolve across both temporal and structured data modalities.}
  \label{fig:tabular}
  \end{center}
  \clearpage
}

\subsection{AIBuildAI delivers strong performance on temporal and tabular tasks}

Temporal and signal modeling focuses on learning predictive patterns from ordered data such as time series and audio signals \cite{oord2016wavenet,Hershey2016CNNAF}, commonly addressed using recurrent neural networks such as LSTMs~\cite{Hochreiter1997lstm}, temporal convolutional networks~\cite{Bai2018AnEE}, WaveNet~\cite{oord2016wavenet}, or spectrogram-based approaches that convert signals to time-frequency representations for processing by image classification models~\cite{Hershey2016CNNAF}. Structured data modeling concerns predictive learning from tabular datasets composed of heterogeneous numerical and categorical variables \cite{grinsztajn2022why}, frequently addressed using gradient-boosted decision trees such as XGBoost~\cite{chen2016XGBoost}, LightGBM~\cite{Ke2017LightGBM}, and CatBoost~\cite{prokhorenkova2018catboost}, as well as neural network approaches such as TabNet~\cite{arik2021tabnet}. We evaluate AIBuildAI on 3 temporal and signal modeling tasks, including scientific signal prediction (seti-breakthrough-listen, icecube-neutrinos-in-deep-ice) and time-series classification (hms-harmful-brain-activity-classification), as well as 3 structured data modeling tasks, including structured regression (new-york-city-taxi-fare-prediction), materials property prediction (nomad2018-predict-transparent-conductors), and general tabular prediction (tabular-playground-series-dec-2021). As shown in Fig.~\ref{fig:tabular}c, AIBuildAI outperforms both baselines (MLEvolve and AIRA-dojo) on all 6 tasks, demonstrating its effectiveness across both temporal and structured data modalities.

Fig.~\ref{fig:tabular}a illustrates an example workflow of AIBuildAI on a temporal signal classification task: hms-harmful-brain-activity-classification. It requires classifying EEG~\cite{jasper1958tentwenty} recordings into six categories of harmful brain activity (e.g., seizure, lateralized periodic discharges). To address this task, the AIBuildAI manager agent invokes the designer sub-agent to propose six solution candidates based on diverse approaches including EfficientNet~\cite{tan2019efficientnet} spectrogram classifiers, WaveNet~\cite{oord2016wavenet} on raw EEG signals, ResNet~\cite{he2016deep} and TinyViT~\cite{wu2022tinyvit} variants, as well as dual-branch hybrids. The manager reviews the six plans, selects a subset of three candidates to proceed, and invokes the coder and tuner sub-agents to implement and evaluate them. Based on the preliminary results, the manager invokes the designer sub-agent to revise candidates 1 and 2 by adding an EEG branch using double banana montage~\cite{acharya2016acns} and a 1D-CNN branch for raw EEG signals, respectively. The coder sub-agent implements the revised designs and the tuner sub-agent retrains the models with adjusted learning rates. The aggregator agent selects the best checkpoint from candidate 1 as the final model.

Fig.~\ref{fig:tabular}b depicts the final solution of AIBuildAI on the hms-harmful-brain-activity-classification task in detail. The pipeline takes two inputs, a spectrogram parquet file and a raw EEG parquet file, and processes them through a dual-branch architecture. The spectrogram branch extracts four EEG regions (e.g., left lateral, right lateral), fills NaN values by zero and clips, applies a log1p transform with Z-score normalization, and resizes the result via bilinear interpolation before feeding it into an ImageNet-pretrained~\cite{deng2009imagenet} EfficientNet-B0~\cite{tan2019efficientnet} with 4-channel input. The EEG branch applies a double banana montage~\cite{acharya2016acns}, Butterworth bandpass filtering~\cite{butterworth1930filter} and clipping, then reshapes and duplicates the signal into 3 channels and resizes via bilinear interpolation before passing it through a separate EfficientNet-B0 with 3-channel input. A fusion MLP head combines the feature representations from both branches to produce 6-class probabilities over the six brain activity categories.

\section{Discussion}\label{Discussion}

The strong performance of AIBuildAI can be attributed to its sub-agent architecture for iteratively refining multiple solutions in parallel. Refining a candidate solution involves heterogeneous operations, such as designing modeling strategies, writing correct code, and training models to achieve optimal performance, each of which requires multi-step reasoning through iterative LLM calls and tool interactions. For example, producing correct code typically requires repeatedly writing code, executing it, inspecting errors, and refining the implementation. Combining all these operations together leads to extensive context lengths that are difficult for a single agent to handle. In contrast, AIBuildAI uses sub-agents for these operations, leveraging the property that one operation does not need to know the details of other operations. For example, a coder sub-agent maintains a coding-focused working context that tracks prior attempts and avoids repeating mistakes, while the tuner sub-agent only needs the final correct code from the coder and its own context focused on optimizing training performance. By decomposing solution refinement into specialized sub-agents, AIBuildAI keeps each agent focused on a well-defined scope, avoiding the context explosion that would arise from maintaining all intermediate reasoning steps in a single agent.

AIBuildAI also differs from prior LLM-based automated AI model development approaches such as AIRA and MLEvolve, which formulate this task as a search over code solutions. In these systems, each modification to a candidate solution is implemented through a single LLM call that directly transforms one code repository into another. As a result, system performance is critically limited by what can be accomplished in a single LLM call, constraining the complexity of modifications at each step. In contrast, AIBuildAI instantiates each operation as a specialized sub-agent rather than a single LLM call. Each sub-agent (designer, coder, tuner) is itself an LLM-based agent with its own prompt, context, and tool-use capabilities, and may execute multiple rounds of reasoning and tool interaction through iterative LLM calls within a single agent invocation. This substantially increases the expressiveness and effectiveness of each operation: a sub-agent can conduct multi-step reasoning and iteratively use tools, whereas a single LLM call must produce a complete modification in one shot. This shift from single-call transformations to multi-step, agent-driven operations is a key factor in AIBuildAI's ability to produce high-quality solutions across diverse tasks.

Despite these advantages, a limitation of AIBuildAI is that the use of multiple interacting sub-agents increases token consumption costs compared with baseline approaches, as each sub-agent inherently conducts multiple rounds of LLM calls. In addition, repeated communication is required across the manager and sub-agent hierarchy, further increasing the cost. AIBuildAI currently uses a single high-capability language model for all sub-agents, but not all LLM calls require the same level of capability. A promising direction is dynamic LLM routing~\cite{ding2024hybridllm,wang2025mixllm}, in which simpler requests (e.g., routine code modifications or configuration updates) are delegated to smaller, cheaper models while only the most challenging operations (e.g., novel architecture design or complex debugging) invoke the most capable model. Such routing mechanisms could potentially reduce costs without sacrificing performance.

Several directions for future work emerge from this study. First, the current evaluation follows the MLE-Bench protocol of a single GPU with a 24-hour time budget, which provides a standardized and reproducible setting. However, real-world AI development tasks, particularly in the era of large language models, often require substantially more computational resources, including multi-GPU training, distributed data parallelism, and longer training horizons spanning days or weeks. As scaling laws have shown that training compute requirements for state-of-the-art models grow rapidly with model and dataset size~\cite{hoffmann2022chinchilla}, extending AIBuildAI to reason about resource allocation becomes increasingly important. For example, adapting modeling strategies based on available compute, automatically selecting distributed training configurations such as parallelism strategies~\cite{zheng2022alpa,zhu2025mist}, or dynamically adjusting time budgets, would improve its applicability to large-scale, production-grade AI development. Second, AIBuildAI currently relies entirely on the parametric knowledge embedded in the underlying language model and a general-purpose web search tool. Incorporating a structured AI task knowledge base, organized as a retrievable collection of modeling strategies, preprocessing techniques, and architecture patterns extracted from the literature and prior successful runs, could improve both the quality and consistency of the system's decisions~\cite{wang2024emgrag}. Such a knowledge base could be organized hierarchically so that sub-agents can efficiently access domain-relevant knowledge, and could be dynamically updated after each run with newly discovered strategies and their empirical outcomes, enabling the system to accumulate experience and improve over time~\cite{xu2025amem}.

More broadly, systems like AIBuildAI could reshape how domain scientists interact with AI across the physical and life sciences, where data are increasingly abundant but AI expertise remains scarce. In genomics, single-cell and spatial transcriptomics experiments now profile thousands of genes across millions of cells, yet building the deep learning models needed for cell-type annotation or perturbation-response prediction still requires specialized bioinformatics skills that most experimental biologists lack~\cite{peidli2024scperturb}. In materials science and chemistry, self-driving laboratories can generate thousands of experiments autonomously~\cite{abolhasani2023selfdrivinglab}, but the surrogate models that guide experimental design must be manually constructed and tuned by dedicated ML engineers. In drug discovery, virtual screening of compound libraries against protein targets demands custom scoring functions and molecular representations that require computational chemistry expertise to build and validate~\cite{sadybekov2023computational}. By automating the full AI model development pipeline, from strategy design through implementation to training and optimization, systems like AIBuildAI could enable scientists in these domains to rapidly build high-performing predictive models directly from their experimental data, accelerating the extraction of scientific insights without requiring deep AI engineering expertise. As experimental throughput continues to outpace analytical capacity across the sciences, the ability to automatically transform raw data into trained, deployable AI models may become as transformative for scientific discovery as the instruments that generate the data themselves.

\section{Methods}
\subsection{LLM-based agent abstraction}

A large language model (LLM) generates outputs conditioned on an input prompt. In a standard single-shot usage, the model is invoked once and its output is returned as the final result. An LLM-based agent extends this paradigm by allowing the model to iteratively interact with an external environment through tool use. Instead of producing only a final answer, the model may generate intermediate outputs that request the execution of external tools (e.g., running code, querying a database, or modifying files). The results of these tool executions are then appended to the context, enabling the model to produce the final answer based on additional information from tool calls.

Formally, an agent is defined by an agent-specific prompt $p_a$ and a set of available tools. Given a user input $u$, the agent first invokes the language model to produce an initial output $y_1 = p_\theta(p_a, u)$. A control loop then examines $y_1$: if it contains a tool invocation, the environment executes the requested tool and returns the tool execution result (observation) $o_1$; otherwise the agent terminates and returns $y_1$ as the final output. If there is a tool invocation, the observation is appended to the context of the agent and the model is invoked again to produce $y_2 = p_\theta(p_a, u, y_1, o_1)$. In general, at step $t$ the model generates $y_t = p_\theta(p_a, u, y_1, o_1, \dots, y_{t-1}, o_{t-1})$, conditioned on the prompt and the full interaction history. Similarly, if $y_t$ contains a tool invocation, the resulting observation $o_t$ is appended to the history. The loop continues until the model produces a response without a tool invocation, or a maximum number of steps is reached. Under this abstraction, a specific agent is fully specified by (i) its prompt $p_a$, which describes the task and how the model should act when intermediate results are presented to the model, and (ii) its tool set, which defines the tools the agent can invoke. By varying these two components with the same underlying language model, one can construct agents with different capabilities and execution strategies.

\subsection{AIBuildAI system design}

We consider the task of automated AI model development, where the goal is to automatically construct a complete solution for a given AI task. The input consists of (i) a natural language description of the AI task and its evaluation protocol, and (ii) a dataset. The desired output is a runnable solution repository, including trained model checkpoints and an inference script that produces predictions for the task on unseen data.

Our approach shares the high-level idea of multi-start local search~\cite{marti2013multistart}. We initialize $n$ solution repositories, denoted $\{\text{repo}_i\}_{i=1}^n$, each representing a fundamentally different modeling strategy for solving the task. Each repository provides an isolated workspace initialized with a copy of the input data, within which a candidate solution can be independently developed and trained. Once initialized, the solution within each repository is iteratively refined: analogous to local search, improvements are incremental rather than fundamental redesigns. At termination, the best-performing solutions are selected or ensembled to produce the final output. Each repository $\text{repo}_i$ maintains a structured collection of artifacts: $\text{data}_i$ denotes a copy of the training data, $\text{plan}_i$ denotes the solution strategy in text, $\text{code}_i$ denotes the implementation of the model and training loop, $\text{config}_i$ contains hyperparameters and training configurations, and $\text{result}_i$ contains execution outputs including training loss, validation performance, and model checkpoints.

To carry out this strategy, we instantiate AIBuildAI as an LLM-based agent following the abstraction described above, with repository-scoped tool variants $\text{read}_i$, $\text{write}_i$, and $\text{execute}_i$ for file manipulation and program execution within each $\text{repo}_i$, as well as a $\text{search}$ tool for retrieving external references on the Internet. The agent iteratively refines each solution trajectory by reading artifacts, modifying plans, code, or configurations, and executing training runs. In principle, a single agent with access to all tools across all repositories could carry out this process. However, refining $n$ parallel solutions involves heterogeneous operations (designing strategies, writing and debugging code, adjusting hyperparameters) across independent repositories, and maintaining all of this information in a single context would rapidly exceed practical context limits and dilute the agent's attention with unrelated details from different methods and repositories.

To address this, AIBuildAI adopts a hierarchical design in which the primary agent (manager) delegates complex operations to specialized sub-agents. Concretely, sub-agents are exposed as tools in the manager's tool set and are invoked by the manager to perform a specific type of refinement on a candidate solution. AIBuildAI includes sub-agents for solution design (designer sub-agent), code generation (coder sub-agent), and model training and hyperparameter tuning (tuner sub-agent), which are detailed in the following subsections. By delegating detailed reasoning and execution to these sub-agents, the manager operates with a compact context that is essential for effectively maintaining a unified control flow across an entire AIBuildAI run. The sub-agents are themselves implemented as LLM-based agents with their own prompts and restricted subsets of the overall tool interface. When invoked, a sub-agent maintains its own internal context with sequences of LLM calls and tool execution results, and writes the resulting artifacts to its repository. This design enables decomposition of the overall task into modular components, enabling more focused and effective behavior.

\subsection{Manager}

The manager is a single top-level agent that orchestrates the overall solution process across all $n$ repositories by selecting and invoking tools based on the current context. Its tool set consists of repository-scoped sub-agent tools ($\text{designer}_i$, $\text{coder}_i$, $\text{tuner}_i$) and $\text{read}_i$ tools for $i$ from $1$ to $n$. The sub-agent tools invoke specialized agents that operate on each repository. The manager understands each sub-agent's behavior through its tool description, which specifies that the designer sub-agent generates or revises solution strategies, the coder sub-agent implements or modifies solutions, and the tuner sub-agent executes training and improves model performance, enabling the manager to reason about when to invoke each. The $\text{read}_i$ tools provide the manager with access to artifacts across all repositories, including plans, code, and results. This enables comparing candidate solutions to select promising ones to proceed with, and understanding the current status of a candidate solution to assign the appropriate sub-agent to refine it. Tool invocations are asynchronous: the manager may issue multiple tool calls in parallel across repositories, allowing concurrent exploration and overlapping execution of design, implementation, and training.

The behavior of the manager is defined by its prompt, which specifies the AI task, the available tools, and the structure of the repository space, and thereby determines how actions are selected during execution. At each step, the manager observes artifacts (via $\text{read}_i$) and selects actions on specific repositories. In typical executions, it initializes multiple repositories by invoking $\text{designer}_i$ to propose candidate strategies, followed by $\text{coder}_i$ and $\text{tuner}_i$ to implement and evaluate them. Subsequent decisions are made adaptively based on repository-specific results. For example, if a solution underperforms due to flaws in the overall approach, the manager may invoke the designer sub-agent to revise the strategy; if implementation issues are detected, it may call the coder sub-agent to modify the code; if performance is limited by training configuration, it may repeatedly invoke the tuner sub-agent to refine hyperparameters. Promising repositories are further refined through such targeted updates, while underperforming ones or similar ones are discarded by ceasing further actions. This flexible, feedback-driven control contrasts with fixed workflows, enabling the manager to dynamically compose different sequences of sub-agent invocations depending on the needs of each candidate solution.

\subsection{Designer}

For each of the $n$ repositories, a separate designer sub-agent is instantiated. Here we describe the $i$-th designer, which is invoked by the manager through the $\text{designer}_i$ tool to operate on $\text{repo}_i$. Its tool set contains $\text{read}_i$ for accessing artifacts in $\text{repo}_i$, $\text{write}_i$ for creating or modifying files, and $\text{search}$ for retrieving external references and best practices. At each invocation, the designer observes the current repository contents via $\text{read}_i$ and produces an updated plan $\text{plan}_i$, adapting its behavior based on the state of the repository. If $\text{plan}_i$ is absent, it generates an initial strategy $\text{plan}_i$ that specifies the modeling approach, data processing pipeline, and training procedure, usually with the help of the web search tool. If $\text{plan}_i$ and $\text{result}_i$ are available, it performs a revision step by analyzing $\text{result}_i$ (e.g., training loss, validation metrics, and logs) to identify potential failure modes such as underfitting, overfitting, or optimization instability, and updates $\text{plan}_i$ accordingly, for example by modifying model architectures, adjusting data processing pipelines, or changing training procedures. The updated plan is written back to $\text{repo}_i$ and subsequently consumed by the coder sub-agent for implementation.

\subsection{Coder}

For each of the $n$ repositories, a separate coder sub-agent is responsible for implementing a solution plan. Here we describe the $i$-th coder, which is invoked by the manager through the $\text{coder}_i$ tool to operate on $\text{repo}_i$. Its tool set contains $\text{read}_i$ for retrieving artifacts, $\text{write}_i$ for creating or modifying files, and $\text{execute}_i$ for executing Python programs in the repository workspace. The coder has no access to other repositories. At each invocation, the coder observes the current repository contents via $\text{read}_i$, in particular $\text{plan}_i$, and produces or updates the implementation $\text{code}_i$, adapting its behavior based on the state of the repository. If $\text{code}_i$ is absent, it generates an initial implementation from $\text{plan}_i$, including the model, training and inference code, and an example configuration file $\text{config}_i$. If $\text{code}_i$ already exists, it performs a revision step to fix errors in $\text{code}_i$, improve robustness, or incorporate updates to $\text{plan}_i$. After implementation, the coder invokes $\text{execute}_i$ to perform a short verification run, ensuring that the pipeline executes end-to-end correctly. If errors are encountered, it conducts iterative debugging until the code is bug-free. The resulting correct code is written to $\text{repo}_i$ for subsequent use by other sub-agents, especially the tuner.

\subsection{Tuner}

For each of the $n$ repositories, a separate tuner sub-agent is responsible for training models and improving performance. Here we describe the $i$-th tuner, which is invoked by the manager through the $\text{tuner}_i$ tool to operate on $\text{repo}_i$. Its tool set contains $\text{read}_i$ for retrieving artifacts, $\text{write}_i$ for modifying configuration files $\text{config}_i$, and $\text{execute}_i$ for executing training runs in the repository workspace. The tuner has no access to other repositories. At each invocation, the tuner observes the current repository contents via $\text{read}_i$, in particular $\text{code}_i$ and $\text{result}_i$, and produces updated results $\text{result}_i$ by changing configuration files based on prior execution outcomes and running the code. Given existing $\text{result}_i$ (including training logs, validation metrics, and checkpoints), it analyzes learning behavior to identify issues such as slow convergence, overfitting, or instability. It then updates training configurations (e.g., hyperparameters, schedules, or data-related settings) and launches new runs via $\text{execute}_i$. As the manager assigns the tuner a maximum tuning time, it uses short preliminary runs to obtain early performance signals. If performance is stagnant or degrades, ineffective configurations are replaced early. In later stages, the tuner extends promising configurations to full training to achieve optimal results within the time budget. All updated configurations and $\text{result}_i$ (logs, checkpoints, and evaluation outputs) are written back to $\text{repo}_i$, delivering validation results to other agents.

\subsection{Setup and aggregation}

In addition to the manager-controlled workflow, AIBuildAI includes two auxiliary sub-agents that are each invoked once: a setup agent at initialization and an aggregator at termination. These agents are not part of the iterative decision loop, but provide deterministic preparation and finalization of the complete workflow.

\paragraph{Setup} The setup agent initializes the model training environment before execution begins. Given the task description, it creates a conda environment and installs the packages required by the subsequent model training process, using conda or pip as appropriate. This ensures that all dependencies are available before any sub-agent begins execution.

\paragraph{Aggregator} The aggregator finalizes the run by producing a single submission from the repositories generated during execution. It operates in a dedicated repository $\text{repo}_{n+1}$, separate from all candidate repositories, to avoid modifying any intermediate artifacts. Its tool set contains $\text{read}_i$ for accessing artifacts across all candidate repositories, and $\text{write}_{n+1}$ and $\text{execute}_{n+1}$ for producing the final submission artifact and executing ensemble inference scripts within $\text{repo}_{n+1}$. The behavior of the aggregator is defined by its prompt $p_{\text{agg}}$, which specifies the objective of maximizing final performance by combining or selecting candidate solutions.

At invocation, the aggregator inspects $\{\text{result}_i\}_{i=1}^n$ and determines whether to select a single best candidate or construct a weighted ensemble, based on the diversity of candidate solutions, their individual validation performance, and the remaining computational budget. When candidates are sufficiently diverse and complementary, the aggregator constructs an ensemble by assigning combination weights derived from validation scores. The specific aggregation strategy is adapted to the task: for classification and detection tasks, the aggregator may apply rank-based blending, in which per-model predictions are converted to rank percentiles before weighted averaging; for segmentation tasks, it may combine predicted probability maps with task-tuned weights and apply binary thresholding; for regression or ranking tasks, it may compute weighted averages of raw predictions. When the remaining time budget is insufficient to execute ensemble inference, or when a single candidate substantially outperforms all others, the aggregator instead selects the best-performing repository and directly retrieves its prediction outputs. In all cases, the aggregator does not retrain models or modify intermediate artifacts, ensuring that the final result is deterministically derived from the executed workflow.

\subsection{Baseline methods}

We compare AIBuildAI with representative approaches for autonomous machine-learning engineering that formulate the problem as a search over code-based solutions. These methods are conceptually related to recent LLM-driven program optimization frameworks~\cite{FunSearch2023,Novikov2025AlphaEvolveAC}, which iteratively generate, evaluate, and refine candidate programs. Under this formulation, each candidate solution is a code artifact $\text{code}$ that implements a complete training and inference pipeline, and is evaluated by executing it to obtain $\text{result} = \text{execute}(\text{code})$ (e.g., validation performance). The search process constructs a tree over candidate solutions: starting from an initial solution, the system selects a previously evaluated solution $\text{code}_t$ (a parent node) and expands it into multiple new variants (child nodes) by modifying $\text{code}_t$ conditioned on $\text{result}_t$. Each new variant is then evaluated by executing its training pipeline, which is computationally expensive. To efficiently allocate computation, these methods employ search strategies, often Monte Carlo Tree Search (MCTS), to balance exploration of new variants and exploitation of high-performing candidates.

\paragraph{AIRA} AIRA~\cite{toledo2025airesearchagentsmachine} instantiates the expansion step through three core operators, each corresponding to a single LLM call that modifies $\text{code}_t$ conditioned on $\text{result}_t$. A Draft operator generates an initial $\text{code}_t$ when no previous solution exists; if $\text{result}_t$ contains an execution error, a Debug operator revises $\text{code}_t$ to produce a corrected variant; if $\text{result}_t$ is valid but performance is unsatisfying, an Improve operator adjusts hyperparameters, model architectures, or feature engineering in $\text{code}_t$. A global Memory operator maintains a running summary of all previous solutions and their results, which is appended as context to every Draft and Improve call. The default search policy is greedy: the system first invokes Draft $n_d$ times to seed an initial population, then at each subsequent step selects the highest-scoring $\text{code}_t$ and applies Improve if it is valid, or Debug otherwise. AIRA also explores alternative search policies, including MCTS and evolutionary methods, though its analysis emphasizes that system effectiveness is primarily determined by the operator set rather than the choice of search algorithm.

\paragraph{MLEvolve} MLEvolve~\cite{Du2025AutoMLGenNF} shares the same core operators as AIRA (Draft, Debug, Improve), each implemented as a single LLM call, but differs in three key aspects. First, it replaces the tree-structured search with Monte Carlo Graph Search (MCGS) by introducing a multi-branch aggregation operator that combines top trajectories from multiple branches to seed entirely new solutions from the root, forming a graph structure that enables information flow beyond standard parent-child relationships while keeping selection and backpropagation on the tree backbone. Second, it removes the global Memory operator and instead introduces two reference-based operators: an intra-branch evolution operator that supplies the nearest $k$ ancestor nodes from the same branch as context, enabling the LLM to reflect on past successes and failures within a single line of investigation; and a cross-branch reference operator that provides top-performing nodes from other branches as context when a branch stagnates, allowing the LLM to incorporate strategies discovered elsewhere. Third, MLEvolve incorporates a domain-specific knowledge base, curated from open-source repositories and competition platforms, organized along model-level (architecture selection), data-level (preprocessing and feature engineering), and strategy-level (test-time augmentation, ensembling) dimensions. Model-level knowledge is optionally injected during initial Draft generation, while data- and strategy-level knowledge is provided as context during subsequent expansion calls.

\subsection{Experimental settings}

All experiments follow the MLE-Bench evaluation protocol~\cite{chan2025mlebench}. AIBuildAI uses Claude Opus 4.6~\cite{anthropic2026claude46} as the backbone LLM for all agents, with a sampling temperature of 1.0 and no nucleus or top-$k$ truncation, and initializes $n=7$ parallel solution repositories per task. It is evaluated with a budget of 24 hours on a Linux x86-64 machine with 24 vCPUs, 256\,GB of RAM, and one NVIDIA A100 GPU. AIRA~\cite{toledo2025airesearchagentsmachine} uses OpenAI o3~\cite{openai2025o3} as its backbone LLM and is evaluated with a budget of 24 hours on a machine with 24 vCPUs, 120\,GB of RAM, and one NVIDIA H200 GPU. MLEvolve~\cite{Du2025AutoMLGenNF} uses Gemini 3 Pro Preview~\cite{google2025gemini3} as its backbone LLM and is evaluated with a budget of 12 hours on a machine with 21 vCPUs, 234\,GB of RAM, and one NVIDIA H200 GPU. 

\section*{Declarations}

\paragraph{Data availability} The benchmark data used in this study are publicly available from the MLE-Bench repository at \url{https://github.com/openai/mle-bench}. The solutions generated by AIBuildAI for the four tasks highlighted in the Results section (alaska2-image-steganalysis, contrails-identify, billion-word-imputation, and hms-harmful-brain-activity-classification) are available at \url{https://drive.google.com/drive/folders/13wgzCp7h0fhWBYrYISxBJWfdPeT8cEK-?usp=sharing}.

\paragraph{Code availability} The AIBuildAI software is publicly available at \url{https://github.com/aibuildai/AI-Build-AI}, together with detailed instructions for using the system.

\paragraph{Author contributions} R.Z., P.Q., Q.C., and L.Z. contributed to conceptualization, methodology, software, investigation, analysis, writing---original draft, and writing---review and editing. P.X. contributed to conceptualization, methodology, investigation, analysis, writing---original draft, and writing---review and editing.

\paragraph{Competing interests} The authors declare no competing interests.


\bibliography{sn-bibliography}

\end{document}